\documentclass[nonblindrev]{informs3} 

\usepackage{balance}
\usepackage{amsmath}
\usepackage[tight,footnotesize]{subfigure}
\usepackage{graphicx}
\usepackage{bm}
\usepackage{algorithm}
\usepackage[noend]{algorithmic}
\usepackage{multirow}
\usepackage{slashbox}
\usepackage{caption}
\usepackage{amsfonts}
\usepackage{siunitx}

\OneAndAHalfSpacedXII 

\usepackage{natbib}
 \bibpunct[, ]{(}{)}{,}{a}{}{,}%
 \def\bibfont{\small}%
\TheoremsNumberedThrough     

\EquationsNumberedThrough    

\MANUSCRIPTNO{2015} 

\begin{document}


\RUNAUTHOR{Tang et al.}

\RUNTITLE{Partial-Adaptive Submodular Maximization}

\TITLE{Partial-Adaptive Submodular Maximization}

\ARTICLEAUTHORS{%
\AUTHOR{Shaojie Tang}
\AFF{Naveen Jindal School of Management, The University of Texas at Dallas}
\AUTHOR{Jing Yuan}
\AFF{Department of Computer Science, The University of Texas at Dallas}
} 

\ABSTRACT{The goal of a typical adaptive sequential decision making problem is to design an interactive policy that selects a group of items sequentially, based on some partial observations, to maximize the expected utility. It has been shown that the utility functions of many real-world applications, including pooled-based active learning and adaptive influence maximization, satisfy the property of adaptive submodularity. However, most of existing studies on adaptive submodular maximization focus on the fully adaptive setting, i.e., one must wait for the feedback from \emph{all} past selections before making the next selection. Although this approach can take full advantage of feedback from the past to make informed decisions, it may take a longer time to complete the selection process as compared with the non-adaptive solution where all selections are made in advance before any observations take place. In this paper, we explore the problem of partial-adaptive submodular maximization where one is allowed to make multiple selections in a batch simultaneously and observe their realizations together. Our approach enjoys the benefits of adaptivity while reducing the time spent on waiting for the observations from past selections. To the best of our knowledge, no results are known for partial-adaptive policies for the non-monotone adaptive
 submodular maximization problem. We study this problem under both cardinality constraint and knapsack constraints, and develop effective and efficient solutions for both cases. We also analyze the batch query complexity, i.e., the number of batches a policy takes to complete the selection process, of our policy under some additional assumptions.}


\maketitle

\section{Introduction}
Adaptive sequential decision making, where one adaptively makes
a sequence of selections based on the stochastic observations collected from the past selections, is at the heart of many machine learning and artificial intelligence tasks. For example,  in experimental design, a practitioner aims to perform a series of tests  in order maximize the amount of ``information'' that can be obtained to  yield valid and objective conclusions. It has been shown that in many real-world applications, including pool-based active learning \cite{golovin2011adaptive}, sensor selection \cite{asadpour2016maximizing}, and adaptive viral marketing \cite{tang2020influence}, the utility function is adaptive submodular. Adaptive submodularity a notion that generalizes the notion
of submodularity from sets to policies. The goal of adaptive submodular maximization is to design an interactive policy that adaptively selects a group of items, where each selection is based on the feedback from the past selections, to maximize an adaptive submodular function subject to some practical constraints. Although this problem has been extensively studied in the literature, most of existing studies focus on the fully adaptive setting where every selection must be made after observing the feedback from \emph{all} past selections. This fully adaptive approach can take full advantage of feedback from the past to make informed decisions, however, as a tradeoff, it may take a longer time to complete the selection process as compared with the non-adaptive solution where all selections are made in advance before any observations take place. This is especially true when the process of collecting the observations from past selections is time consuming. In this paper, we study the problem of partial-adaptive submodular maximization where one is allowed to make multiple selections simultaneously and observe their realizations together. Our setting generalizes both non-adaptive setting and fully adaptive setting. As compared with the fully adaptive strategy, our approach enjoys the benefits of adaptivity while using fewer number of batches. To the best of our knowledge, no results are known for partial-adaptive policies for the non-monotone adaptive
 submodular maximization problem. We next summarize the main contributions made in this paper.
\begin{itemize}
\item We first study the partial-adaptive submodular maximization problem subject to a cardinality constraint. We develop a partial-adaptive greedy policy that achieves a $\alpha/e$ approximation ratio against the optimal fully adaptive policy where $\alpha$ is the \emph{degree of adaptivity} of our policy. One can balance the  the performance/adaptivity tradeoff through adjusting the value of $\alpha$. In particular, if we set $\alpha=1$, our policy reduces to a non-adaptive policy, and if we set $\alpha=1$, our policy reduces to a fully adaptive policy.
\item For the partial-adaptive submodular maximization problem subject to a knapsack constraint, we develop a sampling based partial-adaptive policy that achieves an approximation ratio of $\frac{\min\{1/2, \alpha\}}{ 6+\min\{1, 2\alpha\}}$  with respect to the optimal fully adaptive policy.
\item We theoretically analyze the batch query complexity of our policy and show that if the utility function is policywise submodular, a stronger assumption than adaptive submodularity, then the above sampling based partial-adaptive policy takes at most  $O(\log n \log \frac{B}{c_{\min}})$ number of batches to achieve a constant approximation ratio where $B$ is the budget constraint and $c_{\min}$ is the cost of the cheapest item. It was worth noting that if we consider a cardinality constraint $k$, then $O(\log n \log \frac{B}{c_{\min}})$ is upper bounded by $O(\log n \log k)$ which is polylogarithmic.
\end{itemize}

\section{Related Work}
Maximizing a submodular function subject to various practical constraints  has been extensively studied in the literature \cite{nemhauser1978analysis,badanidiyuru2014fast,buchbinder2014submodular}. However, the classic notation of submodularity is not capable of capturing the interactive nature of many adaptive sequential decision making problems. Recently, \cite{golovin2011adaptive} extends this notation from sets to policies by proposing the concept of adaptive submodularity. Intuitively, it states that if a function is adaptive submodular, then the expected marginal benefit of an item never increases as we collect more observations from past selections. They show that if a function is adaptive monotone and adaptive submodular, then a simple adaptive greedy policy achieves a $(1-1/e)$ approximation ratio. For maximizing a non-monotone adaptive submodular function, \cite{tang2021beyond} show that a randomized greedy policy achieves a $1/e$ approximation ratio.  Unlike most of existing studies on adaptive submodular maximization which is conducted under the fully adaptive setting, our focus is on partial-adaptive (a.k.a batch-mode) setting.

Recently, there are a few results that are related to  partial-adaptive submodular maximization.  \cite{chen2013near} propose a policy that selects batches of fixed
size $r$, and they show that their policy achieves a bounded approximation ratio compare to the optimal policy which is restricted to selecting batches of fixed size $r$.  However, their approximation ratio becomes arbitrarily bad with respect to the optimal fully adaptive policy. In the context of adaptive viral marketing, \cite{yuan2017no} develop a partial-adaptive seeding policy that achieves a bounded approximation ratio against the optimal fully adaptive seeding policy. Our study is similar to theirs in that both studies introduce a controlling parameter to balance the performance/adaptivility tradeoff. However, their results can not be extended to solve a general adaptive submodular maximization problem. Moreover, unlike the batch-mode setting,  their model allows for observing a partial realization of an item during the selection process. \cite{tang2020influence} extends the above study by developing a new partial-adaptive seeding policy subject to a general knapsack constraint. Their design is general enough to be applied to other adaptive submodular maximization problems. Very recently, \cite{esfandiari2019adaptivity} study the batch-mode monotone adaptive submodular optimization problem and they develop an efficient semi adaptive policy  that achieves an almost tight $1-1/e-\epsilon$ approximation ratio. To the best of our knowledge, all existing studies are focusing on maximizing a monotone adaptive submodular function. We are the first to study the non-monotone partial-adaptive submodular maximization problem subject to both cardinality and knapsack constraints. We also provide a rigorous analysis of the batch query complexity of our policy.

\section{Preliminaries and Problem Formulation}
\subsection{Items and States}
The input of our problem is a set $E$ of $n$ items (e.g., tests in experimental design). Each items $e\in E$ has  a random state $\Phi(e) \in O$ where $O$ is a set of all possible states. Let $\phi(e) \in O$ denote a realization of $\Phi(e)$. Thus, a \emph{realization} $\phi$ is a mapping function that maps items to states: $\phi: E \rightarrow O$. In the example of experimental design, the item $e$ may represent a test, such as the temperature, and
$\Phi(e)$ is the result of the test, such as, $\SI{38}{\celsius}$.  There is a known prior probability distribution $p(\phi) = \Pr(\Phi = \phi)$ over realizations $\phi$. When realizations are independent, the distribution $p$ completely factorizes. However, in many real-world applications such as active learning, the realizations of items may depend on each other. For any subset of items $S\subseteq E$, let $\psi: S\rightarrow O$ denote a \emph{partial realization} and $\mathrm{dom}(\psi)=S$ is called the \emph{domain} of $\psi$. Consider a partial  realization $\psi$ and a realization $\phi$, we say $\phi$ is consistent with $\psi$, denoted $\phi \sim \psi$, if they are equal everywhere in $\mathrm{dom}(\psi)$. Moreover, consider two partial realizations $\psi$ and $\psi'$, we say that $\psi$  is a \emph{subrealization} of  $\psi'$, and denoted by $\psi \subseteq \psi'$, if $\mathrm{dom}(\psi) \subseteq \mathrm{dom}(\psi')$ and they agree everywhere in  $\mathrm{dom}(\psi)$. Let $p(\phi\mid \psi)$ represent the conditional distribution over realizations conditional on  a partial realization $\psi$: $p(\phi\mid \psi) =\Pr[\Phi=\phi\mid \Phi\sim \psi ]$. In addition, there is an additive cost function $c(S) = \sum_{e\in S} c(e)$ for any $S\subseteq E$.

\subsection{Policies and Problem Formulation}
A policy is a function $\pi$ that maps a set of partial realizations  to some distribution of $E$: $\pi: 2^{E\times O} \rightarrow \mathcal{P}(E)$, specifying which item to select next. By following a given policy, we can select items adaptively based on our observations made so far. We next introduce two additional notions related to policies \cite{golovin2011adaptive}.

\begin{definition}[Policy  Concatenation]
Given two policies $\pi$ and $\pi'$,  let $\pi @\pi'$ denote a policy that runs $\pi$ first, and then runs $\pi'$, ignoring the observation obtained from running $\pi$.
\end{definition}

\begin{definition}[Level-$t$-Truncation of a Policy]
Given a policy $\pi$, we define its  level-$t$-truncation $\pi_t$  as a policy that runs $\pi$ until it selects $t$ items.
\end{definition}

There is a utility function $f : 2^{E\times O} \rightarrow \mathbb{R}_{\geq0}$ which is defined over items and states.  Let $E(\pi, \phi)$ denote the subset of items selected by $\pi$ under realization $\phi$. The expected  utility $f_{avg}(\pi)$ of a policy $\pi$ can be written as
\begin{eqnarray}
f_{avg}(\pi)=\mathbb{E}_{\Phi\sim p, \Pi}[f(E(\pi, \Phi), \Phi)]~\nonumber
\end{eqnarray}
 where the expectation is taken over possible realizations and the internal randomness of the policy.

In this paper, we first study the problem of  partial-adaptive submodular maximization subject to a cardinality constraint $k$.

  \[\max_{\pi}\{f_{avg}(\pi)\mid |E(\pi, \phi)|\leq k \mbox{ for all realizations } \phi\}\]

Then we generalize this study to consider a knapsack constraint $B$.

  \[\max_{\pi}\{f_{avg}(\pi)\mid c(E(\pi, \phi))\leq B \mbox{ for all realizations } \phi\}\]
\subsection{Adaptive Submodularity and Adaptive Monotonicity}
We next introduce some additional notations that are used in our proofs.
\begin{definition}[Conditional Expected Marginal Utility of an Item]
\label{def:1}
Given a utility function $f : 2^{E\times O} \rightarrow \mathbb{R}_{\geq0}$, the conditional expected marginal utility $\Delta(e \mid S, \psi)$ of an item $e$ on top of a group of items  $S\subseteq E$ conditional on a partial realization $\psi$ is defined as follows
\begin{eqnarray}
\Delta(e \mid  S, \psi)=\mathbb{E}_{\Phi}[f(S \cup \{e\}, \Phi)-f(S, \Phi)\mid \Phi \sim \psi]
\end{eqnarray}
where the expectation is taken over $\Phi$ with respect to $p(\phi\mid \psi)=\Pr(\Phi=\phi \mid \Phi \sim \psi)$.
\end{definition}

\begin{definition}[Conditional Expected Marginal Utility of a Policy]
\label{def:policy}
Given a utility function $f : 2^{E\times O} \rightarrow \mathbb{R}_{\geq0}$, the conditional expected marginal utility $\Delta(\pi \mid S, \psi)$ of a policy $\pi$ on top of a group of items  $S\subseteq E$ conditional on a partial realization $\psi$ is
\[\Delta(\pi\mid S, \psi)=\mathbb{E}_{\Phi, \Pi}[f(S \cup E(\pi, \Phi), \Phi)-f(S, \Phi)\mid \Phi\sim \psi]\]
where the expectation is taken over $\Phi$ with respect to $p(\phi\mid \psi)=\Pr(\Phi=\phi \mid \Phi \sim \psi)$ and the random output of $\pi$.
\end{definition}

Now we are ready to introduce the notations of adaptive submodularity and adaptive monotone \cite{golovin2011adaptive}. Intuitively, adaptive submodularity is a generalization of the classic notation of submodularity  from sets to policies. This condition states that the expected marginal benefit of an item never increases as we collect more observations from past selections.

\begin{definition}[Adaptive Submodularity and Adaptive Monotonicity]
\label{def:122}A function $f : 2^{E\times O} \rightarrow \mathbb{R}_{\geq0}$ is submodular adaptive if for any two partial realizations $\psi$ and $\psi'$ such that $\psi\subseteq \psi'$, the following holds for each $e\in E\setminus \mathrm{dom}(\psi')$:
\begin{eqnarray}\label{def:22}
\Delta(e\mid \mathrm{dom}(\psi), \psi) \geq \Delta(e\mid \mathrm{dom}(\psi'), \psi')
\end{eqnarray}
Moreover, we say a utility function $f : 2^{E\times O} \rightarrow \mathbb{R}_{\geq0}$  is \emph{adaptive monotone} \citep{golovin2011adaptive} if  for any partial realization $\psi$ and any $e\in E\setminus \mathrm{dom}(\psi)$: $\Delta(e\mid \mathrm{dom}(\psi), \psi) \geq 0$.
\end{definition}

\section{Cardinality Constraint}
  We first study the problem of  partial-adaptive submodular maximization subject to a cardinality constraint $k$. It has been shown that if the utility function is adaptive submodular, then a fully adaptive greedy policy can achieve a $1/e$ approximation ratio against the optimal fully adaptive policy \citep{tang2021beyond}. However, one weakness about a fully adaptive policy is that  one must wait for the observations from all past selections before making a new selection. To this end, we develop a \emph{Partial-Adaptive Greedy Policy} $\pi^{p}$ that allows to make multiple selections simultaneously within a single batch. We show that $\pi^{p}$  achieves a $\alpha/e$ approximation ratio with respect to the optimal fully adaptive policy where $\alpha\in[0,1]$ is called \emph{degree of adaptivity}. One can adjust the value of $\alpha$ to  balance the performance/adaptivity tradeoff. 


\subsection{Algorithm Design}

  We next explain the design of $\pi^{p}$ (a detailed implementation of $\pi^{p}$ is listed in Algorithm \ref{alg:LPP2}).    We first add a set $V$ of $2k-1$ dummy items to the ground set, such that $\Delta(e \mid \mathrm{dom}(\psi), \psi) =0$ for any $e \in V$ and any
partial realization $\psi$. Let $E'=E\cup V$ denote the expanded ground set. These dummy items are added to avoid selecting any item with negative marginal utility. Note that we can safely remove all these dummy items from the solution without affecting its utility. For any partial realization $\psi$ and any subset of items $S\subseteq \mathrm{dom}(\psi)$, define $M(S, \psi)$ as a set of $k$ items that have the largest marginal utility on top of $S$ conditional on $\psi$, i.e.,
\begin{eqnarray}
M(S, \psi) \in  \argmax_{V\subseteq E'; |V|= k} \sum_{e\in E'}\Delta(e\mid S, \psi)
\end{eqnarray}

For any iteration $t\in[k]$ of $\pi^{p}$, let $S_t$ denote the first $t$ items selected by $\pi^{p}$, let $b[t]$ denote the batch index of the $t$-th item selected by $\pi^{p}$, i.e., the $t$-th item is selected in batch $b[t]$, and let $S_{[q]}$ denote the set of selected items from batch $q$. Let $\psi_q$ denote the partial realization of the first $q$ batches of items selected by $\pi^{p}$, i.e., $\mathrm{dom}(\psi_q) = \cup_{i\in[1, q]}{S_{[i]}}$.   Set the initial solution $S_0=\emptyset$ and the initial partial realization $\psi_0=\emptyset$.

\begin{itemize}
\item   Starting with the first iteration $t=1$.
\item In each iteration $t$, we compare $\sum_{e\in M(S_{t-1}, \psi_{b[t-1]-1})}\Delta(e\mid S_{t-1}, \psi_{b[t-1]-1})$ with $\alpha\cdot  \sum_{e\in M(\mathrm{dom}(\psi_{b[t-1]-1}), \psi_{b[t-1]-1})}\Delta(e\mid \mathrm{dom}(\psi_{b[t-1]-1}), \psi_{b[t-1]-1})$, then decide whether to start a new batch or not based on the result of the comparison as follows. 
\begin{itemize}
\item If
\begin{eqnarray}
\label{eq:ooop}
\sum_{e\in M(S_{t-1}, \psi_{b[t-1]-1})}\Delta(e\mid S_{t-1}, \psi_{b[t-1]-1}) \geq \alpha\cdot  \sum_{e\in M(\mathrm{dom}(\psi_{b[t-1]-1}), \psi_{b[t-1]-1})}\Delta(e\mid \mathrm{dom}(\psi_{b[t-1]-1}), \psi_{b[t-1]-1}),
\end{eqnarray}
then $\pi^{p}$ chooses to stay with the current batch, i.e., $b[t]=b[t-1]$. It  samples an
item $e_t$  uniformly at random  from  $M(S_{t-1}, \psi_{b[t]-1})$, which is identical to $M(S_{t-1}, \psi_{b[t-1]-1})$ due to  $b[t]=b[t-1]$,  and updates the solution $S_t$ using $S_{t-1}\cup\{e_t\}$. Move to the next iteration $t=t+1$.
\item Otherwise, $\pi^{p}$ starts a new batch, i.e., $b[t]=b[t-1]+1$, and observe the partial realization $\Phi(e)$ of all items $e$ from the previous batch $S_{[b[t-1]]}$. Then it updates the observation $\psi_{b[t]-1}$ using $\psi_{b[t-1]-1}\cup \{(e, \Phi(e)) \mid e\in S_{[b[t-1]]}\}$. Note that $S_{t-1}=\mathrm{dom}(\psi_{b[t]-1})$ in this case. At last, it samples an item  $e_t$ uniformly at random  from  $M(S_{t-1}, \psi_{b[t]-1})$ and  updates the solution $S_t$ using $S_{t-1}\cup\{e_t\}$. Move to the next iteration $t=t+1$.
\end{itemize}
\item The above process iterates until  $\pi^{p}$ selects $k$ items (which may include some dummy items).
\end{itemize}

\emph{Remark:} Note that $\sum_{e\in M(\mathrm{dom}(\psi_{b[t-1]-1}), \psi_{b[t-1]-1})}\Delta(e\mid \mathrm{dom}(\psi_{b[t-1]-1}), \psi_{b[t-1]-1})$ in (\ref{eq:ooop}) is an upper bound of $\sum_{e\in M(S_{t-1}, \psi_{b[t-1]-1})}\Delta(e\mid S_{t-1}, \psi_{b[t-1]-1})$ due to $\mathrm{dom}(\psi_{b[t-1]-1})\subseteq S_{t-1}$ and $f : 2^{E\times O} \rightarrow \mathbb{R}_{\geq0}$ is adaptive submodular. Intuitively, satisfying (\ref{eq:ooop}) ensures that the expected gain of each iteration is sufficiently large to achieve a constant approximation ratio. Unlike some other criteria proposed in previous studies \citep{tang2020influence,esfandiari2019adaptivity}, evaluating (\ref{eq:ooop}) is relatively easy since it does not involve the calculation of the expectation of the maximum of $n$ random variables. Under our framework, one can adjust the degree of adaptivity $\alpha\in[0,1]$ to balance the performance/adaptivity tradeoff. In particular, choosing a smaller $\alpha$ makes it easier to satisfy (\ref{eq:ooop}) and hence leads to fewer number of batches but poorer performance. Consider an extreme case when $\alpha=0$, $\pi^{p}$ takes only one batch to complete the selection process, i.e., $\pi^{p}$ reduces to a non-adaptive policy in this case. At the other extreme, if we set $\alpha=1$, then our policy becomes fully adaptive.
\begin{algorithm}[hptb]
\caption{Partial-Adaptive Greedy Policy $\pi^p$}
\label{alg:LPP2}
\begin{algorithmic}[1]
\STATE $t=1; b[0]=1; \psi_0=\emptyset; S_0=\emptyset; \forall i\in[n], S_{[i]}=\emptyset$.
\WHILE {$t\leq k$}
\STATE let $M(S_{t-1}, \psi_{b[t-1]-1}) \leftarrow \argmax_{V\subseteq E'; |V|= k} \sum_{e\in E'}\Delta(e\mid S_{t-1}, \psi_{b[t-1]-1})$;
\IF {$\sum_{e\in M(S_{t-1}, \psi_{b[t-1]-1})}\Delta(e\mid S_{t-1}, \psi_{b[t-1]-1}) \geq \alpha\cdot  \sum_{e\in M(\mathrm{dom}(\psi_{b[t-1]-1}), \psi_{b[t-1]-1})}\Delta(e\mid \mathrm{dom}(\psi_{b[t-1]-1}), \psi_{b[t-1]-1})$}
\STATE \COMMENT{stay in the current batch}
\STATE $b[t]=b[t-1]$;
\STATE sample $e_t$ uniformly at random from $M(S_{t-1}, \psi_{b[t]-1})$;
\STATE $S_t=S_{t-1}\cup\{e_t\}$; $S_{[b[t]]}=S_{[b[t]]}\cup\{e_t\}$;
\ELSE
\STATE \COMMENT{start a new batch}
\STATE $b[t]=b[t-1]+1$;
\STATE observe $\{(e, \Phi(e)) \mid e\in S_{[b[t-1]]}\}$; $\psi_{b[t]-1}=\psi_{b[t-1]-1}\cup \{(e, \Phi(e)) \mid e\in S_{[b[t-1]]}\}$;
\STATE let $M(S_{t-1}, \psi_{b[t]-1}) \leftarrow \argmax_{V\subseteq E'; |V|= k} \sum_{e\in E'}\Delta(e\mid S_{t-1}, \psi_{b[t]-1})$;
\STATE sample $e_t$ uniformly at random from $M(S_{t-1}, \psi_{b[t]-1})$; $S_t=S_{t-1}\cup\{e_t\}$; $S_{[b[t]]}=S_{[b[t]]}\cup\{e_t\}$;
\ENDIF
\STATE $t\leftarrow t+1$;
\ENDWHILE
\end{algorithmic}
\end{algorithm}

\subsection{Performance Analysis}
We next analyze the performance of $\pi^{p}$  against the optimal fully adaptive strategy. The following main theorem shows that $\pi^{p}$ with degree of adaptivity $\alpha$  achieves an approximation ratio of $\alpha/e$.
\begin{theorem}
\label{thm:1}
If  $f: 2^{E \times O}\rightarrow \mathbb{R}_{\geq 0}$ is adaptive submodular, then the Partial-Adaptive Greedy Policy $\pi^{p}$ with degree of adaptivity $\alpha$ achieves a $\alpha/e$ approximation ratio in expectation.
\end{theorem}

The rest of this section is devoted to proving Theorem \ref{thm:1}. We first present two technical lemmas. Recall that for any iteration $t\in[k]$, $S_{t-1}$ represents the first $t-1$ selected items, $\psi_{b[t]-1}$ represents the partial realization of all items selected from the first $b[t]-1$ batches, and $M(S, \psi) \in  \argmax_{V\subseteq E'; |V|= k} \sum_{e\in E'}\Delta(e\mid S, \psi)$.
%
%

\begin{lemma}
\label{lem:cocoon}
For  each iteration $t\in[k]$, $\sum_{e\in M(S_{t-1}, \psi_{b[t]-1})} \Delta(e \mid S_{t-1}, \psi_{b[t]-1}) \geq \alpha\cdot \sum_{e\in M(\mathrm{dom}(\psi_{b[t]-1}), \psi_{b[t]-1})} \Delta(e \mid \mathrm{dom}(\psi_{b[t]-1}), \psi_{b[t]-1})$.
\end{lemma}
\emph{Proof:} We prove this lemma in two cases.
\begin{itemize}
\item If $b[t]=b[t-1]$, i.e., the $t-1$-th item and the $t$-th item are selected within the same batch, then according to the design of $\pi^p$, it must be the case that $\sum_{e\in M(S_{t-1}, \psi_{b[t-1]-1})}\Delta(e\mid S_{t-1}, \psi_{b[t-1]-1}) \geq \alpha\cdot  \sum_{e\in M(\mathrm{dom}(\psi_{b[t-1]-1}), \psi_{b[t-1]-1})}\Delta(e\mid \mathrm{dom}(\psi_{b[t-1]-1}), \psi_{b[t-1]-1})$. This together with $b[t]=b[t-1]$ implies that $\sum_{e\in M(S_{t-1}, \psi_{b[t-1]-1})}\Delta(e\mid S_{t-1}, \psi_{b[t-1]-1}) \geq \alpha\cdot  \sum_{e\in M(\mathrm{dom}(\psi_{b[t]-1}), \psi_{b[t]-1})}\Delta(e\mid \mathrm{dom}(\psi_{b[t]-1}), \psi_{b[t]-1})$.
\item Otherwise, if  $b[t]=b[t-1]+1$, i.e.,  the $t-1$-th item and the $t$-th item are selected within different batches, then we have $\mathrm{dom}(\psi_{b[t]-1})=S_{t-1}$. Thus, $\sum_{e\in M(S_{t-1}, \psi_{b[t]-1})} \Delta(e \mid S_{t-1}, \psi_{b[t]-1}) = \sum_{e\in M(\mathrm{dom}(\psi_{b[t]-1}), \psi_{b[t]-1})} \Delta(e \mid \mathrm{dom}(\psi_{b[t]-1}), \psi_{b[t]-1}) \geq \alpha\cdot\sum_{e\in M(\mathrm{dom}(\psi_{b[t]-1}), \psi_{b[t]-1})} \Delta(e \mid \mathrm{dom}(\psi_{b[t]-1}), \psi_{b[t]-1})$ where the inequality is due to $\alpha \leq 1$.
\end{itemize}
This finishes the proof of this lemma. $\Box$

The next lemma shows that for any iteration $t\in[k]$, the sum of expected marginal benefits of all items from $ M(\mathrm{dom}(\psi_{b[t]-1}), \psi_{b[t]-1})$ is sufficiently high. This will be used later to lower bound the expected gain of each iteration of our policy.
\begin{lemma}
\label{lem:isaac}
Let $\pi^*$ denote an optimal fully adaptive policy. In each iteration $t\in[k]$, $\sum_{e\in M(\mathrm{dom}(\psi_{b[t]-1}), \psi_{b[t]-1})} \Delta(e \mid \mathrm{dom}(\psi_{b[t]-1}), \psi_{b[t]-1})\geq \Delta(\pi^*\mid S_{t-1}, \psi_{b[t]-1})$.
\end{lemma}

\emph{Proof:}  Observe that for any $S_{t-1}$ and $\psi_{b[t]-1}$, the marginal utility of the optimal policy $\pi^*$ on top of $S_{t-1}$ conditional on    $\psi_{b[t]-1}$ can be represented as
\[\Delta(\pi^*\mid S_{t-1}, \psi_{b[t]-1}) = \mathbb{E}_{\Phi \sim \psi_{b[t]-1}}[\Delta(\pi^*\mid S_{t-1}, \psi_{b[t]-1} \cup_{e\in S_{t-1} \setminus \mathrm{dom}(\psi_{b[t]-1}) }(e, \Phi(e)))]\]

In the above equation,  $\psi_{b[t]-1} \cup_{e\in S_{t-1} \setminus \mathrm{dom}(\psi_{b[t]-1}) }(e, \Phi(e))$ represents a random realization of $S_{t-1}$ conditional on $\psi_{b[t]-1}$. For convenience, let $\Phi( S_{t-1})$ denote  $\psi_{b[t]-1} \cup_{e\in S_{t-1} \setminus \mathrm{dom}(\psi_{b[t]-1}) }(e, \Phi(e))$ for short. Then we have
\begin{eqnarray}
\Delta(\pi^*\mid S_{t-1}, \psi_{b[t]-1}) &=& \mathbb{E}_{\Phi \sim \psi_{b[t]-1}}[\Delta(\pi^*\mid S_{t-1}, \Phi( S_{t-1}))]\\
&\leq& \mathbb{E}_{\Phi \sim \psi_{b[t]-1}}[\sum_{e\in M(S_{t-1}, \Phi( S_{t-1}))} \Delta(e \mid  S_{t-1}, \Phi( S_{t-1}))]\\
&\leq& \mathbb{E}_{\Phi \sim \psi_{b[t]-1}}[\sum_{e\in M(S_{t-1}, \Phi( S_{t-1}))} \Delta(e \mid \mathrm{dom}(\psi_{b[t]-1}), \psi_{b[t]-1})]\\
&\leq& \mathbb{E}_{\Phi \sim \psi_{b[t]-1}}[\sum_{e\in M(\mathrm{dom}(\psi_{b[t]-1}), \psi_{b[t]-1})} \Delta(e \mid \mathrm{dom}(\psi_{b[t]-1}), \psi_{b[t]-1})]\\
&=& \sum_{e\in M(\mathrm{dom}(\psi_{b[t]-1}), \psi_{b[t]-1})} \Delta(e \mid \mathrm{dom}(\psi_{b[t]-1}), \psi_{b[t]-1})
\end{eqnarray}
The first inequality is due to $f: 2^{E \times O}\rightarrow \mathbb{R}_{\geq 0}$ is adaptive submodular and Lemma 1 in \citep{gotovos2015non}. The second inequality is due to $f: 2^{E \times O}\rightarrow \mathbb{R}_{\geq 0}$ is adaptive submodular and $\psi_{b[t]-1} \subseteq \Phi( S_{t-1})$. $\Box$

Now we are ready to prove the main theorem. Let the random variable $\mathcal{S}_{t-1}$ denote the first $t-1$ items selected by $\pi^p$ and let the random variable $\Psi_{b[t]-1}$ denote the partial realization of the first $b[t]-1$ batches of items selected by $\pi^p$.  For any $t\in[k]$, we next bound the expected marginal gain of the $t$-th iteration of $\pi^p$,
\begin{eqnarray}
f_{avg}(\pi^{p}_t) - f_{avg}(\pi^{p}_{t-1}) &=& \mathbb{E}_{\mathcal{S}_{t-1}, \Psi_{b[t]-1}}[\mathbb{E}_{e_t}[\Delta(e_t \mid\mathcal{S}_{t-1}, \Psi_{b[t]-1})]]\nonumber\\
&=& \frac{1}{k} \mathbb{E}_{\mathcal{S}_{t-1}, \Psi_{b[t]-1}} [\sum_{e\in M(\mathcal{S}_{t-1}, \Psi_{b[t]-1})} \Delta(e \mid\mathcal{S}_{t-1}, \Psi_{b[t]-1})]\nonumber\\
&\geq& \frac{1}{k} \mathbb{E}_{\mathcal{S}_{t-1}, \Psi_{b[t]-1}} [\alpha\cdot \sum_{e\in M(\mathrm{dom}(\Psi_{b[t]-1}), \Psi_{b[t]-1})} \Delta(e \mid \mathrm{dom}(\Psi_{b[t]-1}), \Psi_{b[t]-1})]\nonumber\\
&=& \frac{\alpha}{k} \mathbb{E}_{\mathcal{S}_{t-1}, \Psi_{b[t]-1}} [ \sum_{e\in M(\mathrm{dom}(\Psi_{b[t]-1}), \Psi_{b[t]-1})} \Delta(e \mid \mathrm{dom}(\Psi_{b[t]-1}), \Psi_{b[t]-1})]\nonumber\\
&\geq& \frac{\alpha}{k} \mathbb{E}_{\mathcal{S}_{t-1}, \Psi_{b[t]-1}} [ \Delta(\pi^*\mid\mathcal{S}_{t-1}, \Psi_{b[t]-1})]\nonumber\\
&=&  \frac{\alpha}{k} (f_{avg}(\pi^*@\pi^{p}_{t-1})-  f_{avg}(\pi^{p}_{t-1}))\label{eq:1}\\
&\geq&  \frac{\alpha}{k}((1-\frac{1}{k})^{t-1} f_{avg}(\pi^*)-  f_{avg}(\pi^{p}_{t-1}))\label{eq:bbbb}
\end{eqnarray}
The second equality is due to the fact that at each round $t\in[k]$,  $\pi^p$  adds an
item  uniformly at random  from  $M(S_{t-1}, \psi_{b[t]-1})$ to the solution. The first inequality is due to Lemma \ref{lem:cocoon}. The second inequality is due to Lemma \ref{lem:isaac}. 
The last inequality is due to Lemma 1 in \citep{tang2021beyond} where they show that $f_{avg}(\pi^*@\pi^{p}_{t-1})\geq (1-\frac{1}{k})^{t-1} f_{avg}(\pi^*)$.

 We next prove
 \begin{equation}\label{eq:xxx}
 f_{avg}(\pi^{p}_{t})\geq \frac{\alpha t}{k} (1-\frac{1}{k})^{t-1}f_{avg}(\pi^*)
 \end{equation} by induction on $t$. For $t=0$,  $f_{avg}(\pi^{p}_{0})\geq 0 \geq \frac{\alpha\cdot 0}{k} (1-\frac{1}{k})^{0-1}f_{avg}(\pi^*)$. Assume Eq. (\ref{eq:xxx}) is true for $t'<t$, let us
prove it for $t$.
\begin{eqnarray*}
 f_{avg}(\pi^{p}_{t}) &\geq&  f_{avg}(\pi^{p}_{t-1})+ \frac{\alpha}{k}((1-\frac{1}{k})^{t-1} f_{avg}(\pi^*)-  f_{avg}(\pi^{p}_{t-1}))\\
 &=& (1-\alpha/k)f_{avg}(\pi^{p}_{t-1})+\frac{\alpha(1-\frac{1}{k})^{t-1} f_{avg} (\pi^*)}{k}\\
  &\geq& (1-\alpha/k)\cdot (\alpha(t-1)/k)\cdot (1-1/k)^{t-2}\cdot f_{avg}(\pi^*)+\frac{\alpha(1-\frac{1}{k})^{t-1} f_{avg} (\pi^*)}{k}\\
  &\geq& (1-1/k)\cdot (\alpha(t-1)/k)\cdot (1-1/k)^{t-2}\cdot f_{avg}(\pi^*)+\frac{\alpha(1-\frac{1}{k})^{t-1} f_{avg} (\pi^*)}{k}\\
  &=&  \frac{\alpha t}{k} (1-\frac{1}{k})^{t-1}f_{avg}(\pi^*)
\end{eqnarray*}
The first inequality is due to (\ref{eq:bbbb}), the second inequality is due to the inductive assumption.
When $t=k$, we have $ f_{avg}(\pi^{p}_{t}) \geq \alpha (1-1/k)^{k-1}\cdot f_{avg}(\pi^*) \geq (\alpha/e) f_{avg}(\pi^*)$. This finishes the proof of the main theorem.

\emph{Remark:} When the utility function $f: 2^{E \times O}\rightarrow \mathbb{R}_{\geq 0}$  is adaptive submodular and adaptive monotone, \cite{golovin2011adaptive} show that $f_{avg}(\pi^*@\pi^{p}_{t-1})\geq f_{avg}(\pi^*)$ for all $t\in [k]$. Thus, for all $t\in [k]$, \begin{eqnarray}
f_{avg}(\pi^{p}_t) - f_{avg}(\pi^{p}_{t-1}) &\geq&  \frac{\alpha}{k} (f_{avg}(\pi^*@\pi^{p}_{t-1})-  f_{avg}(\pi^{p}_{t-1}))\nonumber\\
&\geq&  \frac{\alpha}{k}( f_{avg}(\pi^*)-  f_{avg}(\pi^{p}_{t-1}))\label{eq:bbbb1}
\end{eqnarray}
The first inequality is due to (\ref{eq:1}). Through induction on $t$, we have $ f_{avg}(\pi^{p}) \geq (1-e^{-\alpha}) f_{avg}(\pi^*)$.

\begin{theorem}
\label{thm:4}
If $f: 2^{E \times O}\rightarrow \mathbb{R}_{\geq 0}$  is adaptive submodular and adaptive monotone, then the Partial-Adaptive Greedy Policy $\pi^p$ achieves a $1-e^{-\alpha}$ approximation ratio in expectation.
\end{theorem}

\section{Knapsack Constraint}
 In this section, we study our problem subject to a knapsack constraint $B$. In \citep{tang2021pointwise,amanatidis2020fast}, they develop a fully adaptive policy that achieves a bounded approximation ratio against the optimal fully adaptive policy. We extend their design by developing  a partial-adaptive policy which allows to select multiple items in a single batch. Our policy with degree of adaptivity $\alpha$ achieves an approximation ratio of $\frac{1}{6+4/\alpha}$  with respect to the optimal fully adaptive policy. Again, one can balance the peformance/adaptivity tradeoff by adjusting the degree of adaptivity $\alpha$. In addition, we provide a rigorous analysis of the batch query complexity of our policy under some additional assumptions.

\begin{algorithm}[hptb]
\caption{ Partial-Adaptive Density-Greedy Policy  $\pi^{k}$}
\label{alg:LPP2}
\begin{algorithmic}[1]
\STATE $t=1; b[0]=1; \psi_0=\emptyset; S_0=\emptyset; \forall i\in[n], S_{[i]}=\emptyset$.
\FOR {$i\in E$}
\STATE let $r \sim \mathrm{Bernoulli}(\frac{1}{2})$
\IF {$r=1$}
\STATE add $i$ to $R$
\ENDIF
\ENDFOR
\WHILE {$R\setminus S_{t-1} \neq \emptyset$}
\STATE let $e' \leftarrow \arg\max_{e \in R\setminus S_{t-1}} \frac{\Delta(e\mid  S_{t-1}, \psi_{b[t-1]-1})}{c(e)}$
\IF {$t=1$}
\STATE $e''\leftarrow e'$
\ENDIF
\IF {$\frac{\Delta(e'\mid  S_{t-1}, \psi_{b[t-1]-1})}{c(e')}\geq \alpha\cdot \frac{\Delta(e''\mid  \mathrm{dom}(\psi_{b[t-1]-1}), \psi_{b[t-1]-1})}{c(e'')}$}
\STATE \COMMENT{stay in the current batch}
\IF {$c(e')+c(S_{t-1})\leq B$}
\STATE $b[t]=b[t-1]; e_t\leftarrow e'$; $S_t=S_{t-1}\cup\{e_t\}; S_{[b[t]]}=S_{[b[t]]}\cup\{e_t\}; t=t+1$
\ELSE
\STATE  break  \label{line:1}
\ENDIF
\ELSE
\STATE \COMMENT{start a new batch}
\STATE $b[t]=b[t-1]+1$
\STATE observe $\{(e, \Phi(e)) \mid e\in S_{[b[t-1]]}\}$; $\psi_{b[t]-1}=\psi_{b[t-1]-1}\cup\{(e, \Phi(e)) \mid e\in S_{[b[t-1]]}\}$;
\STATE let $e''\leftarrow \arg\max_{e\in  R\setminus \mathrm{dom}(\psi_{b[t]-1}) }\frac{\Delta(e\mid \mathrm{dom}(\psi_{b[t]-1}), \psi_{b[t]-1})}{c(e)}$
\IF {$\frac{\Delta(e'' \mid \mathrm{dom}(\psi_{b[t]-1}), \psi_{b[t]-1})}{c(e'')}>0$}
\IF {$c(e'')+c(S_{t-1})\leq B$}
\STATE $e_t\leftarrow e''$; $S_t=S_{t-1}\cup\{e_t\}$; $S_{[b[t]]}=S_{[b[t]]}\cup\{e_t\}$; $F=F\setminus\{e_t\}; t=t+1$
\ELSE \STATE break \label{line:21} 
\ENDIF
\ELSE
\STATE break \label{line:3}
\ENDIF
\ENDIF
\ENDWHILE
\end{algorithmic}
\end{algorithm}
\subsection{Algorithm Design}
\label{sec:algorithm}
We first construct two candidate policies: the first policy always picks the best singleton $o$ that maximizes the expected utility, i.e., $o \in  \arg\max_{e\in E} \mathbb{E}_{\Phi\sim p}[f(\{e\}, \Phi)]$ and the second candidate is a sampling based ``density-greedy'' policy $\pi^{k}$. Our final policy picks one from the above two candidates to execute such that  $\{o\}$ is picked with probability $\frac{1/\alpha}{3+2/\alpha}$  and $\pi^k$ is picked with probability $\frac{3+ 1/\alpha}{3+2/\alpha}$. In the rest of this paper, let $f(o)$ denote $\mathbb{E}_{\Phi\sim p}[f(\{o\}, \Phi)]$ for short.

We next explain the idea of \emph{Partial-Adaptive Density-Greedy Policy} $\pi^{k}$ (the second candidate policy). $\pi^k$ first selects a random subset $R$ which is obtained by including each item $e\in E$ independently with probability $1/2$. Then we run a ``density-greedy'' algorithm only on $R$. We first introduce some notations. For each iteration $t\in[n]$, let $b[t]$ denote the batch index of $t$, i.e., the $t$-th item selected in batch $b[t]$, for convenience, we define $b[0]=1$. Let $S_{t-1}$ denote the first $t-1$ items selected by $\pi^k$, and $\psi_{b[t-1]-1}$ represent the partial realization of the first $b[t-1]-1$ batches of selected items.  Set the initial solution $S_0=\emptyset$ and the initial partial realization $\psi_0=\emptyset$.

\begin{itemize}
\item Starting from iteration $t=1$ and batch $b[t]=1$.
\item In each iteration $t$, let $e'$ be the item that has the largest benefit-cost ratio on top of $S_{t-1}$ conditional on $\psi_{b[t-1]-1}$ from $R\setminus S_{t-1}$, i.e.,
\begin{eqnarray}
e' \leftarrow \arg\max_{e\in R\setminus S_{t-1}} \frac{\Delta(e\mid  S_{t-1}, \psi_{b[t-1]-1})}{c(e)}
\end{eqnarray}

Let $e'' $ be the item that has the largest benefit-cost ratio on top of $ \mathrm{dom}(\psi_{b[t-1]-1})$ conditional on $\psi_{b[t-1]-1}$ from $R\setminus \mathrm{dom}(\psi_{b[t-1]-1})$, i.e.,
\begin{eqnarray}
e'' \leftarrow \arg\max_{e\in  R\setminus \mathrm{dom}(\psi_{b[t-1]-1}) }\frac{\Delta(e\mid \mathrm{dom}(\psi_{b[t-1]-1}), \psi_{b[t-1]-1})}{c(e)}
\end{eqnarray}
It will become clear later that  $e''$ stores the first selected item, if any, from the $b[t-1]$-th batch. Note that $\mathrm{dom}(\psi_{b[t-1]-1}) \subseteq  S_{t-1}$.

Compare $\frac{\Delta(e'\mid  S_{t-1}, \psi_{b[t-1]-1})}{c(e')}$ with $\alpha\cdot \frac{\Delta(e''\mid  \mathrm{dom}(\psi_{b[t-1]-1}), \psi_{b[t-1]-1})}{c(e'')}$,
\begin{itemize}
\item if $\frac{\Delta(e'\mid  S_{t-1}, \psi_{b[t-1]-1})}{c(e')}\geq \alpha\cdot \frac{\Delta(e''\mid  \mathrm{dom}(\psi_{b[t-1]-1}), \psi_{b[t-1]-1})}{c(e'')}$ and adding $e'$ to the solution does not violate the budget constraint, then stay in the current batch, i.e., $b[t] = b[t-1]$, add $e'$ to the solution, i.e, $S_t=S_{t-1}\cup\{e'\}$. Move to the next iteration, i.e., $t=t+1$;
\item if $\frac{\Delta(e'\mid  S_{t-1}, \psi_{b[t-1]-1})}{c(e')}\geq \alpha\cdot \frac{\Delta(e''\mid  \mathrm{dom}(\psi_{b[t-1]-1}), \psi_{b[t-1]-1})}{c(e'')}$ and adding $e'$ to the solution violates the budget constraint, then terminate;
\item if $\frac{\Delta(e'\mid  S_{t-1}, \psi_{b[t-1]-1})}{c(e')}< \alpha\cdot \frac{\Delta(e''\mid  \mathrm{dom}(\psi_{b[t-1]-1}), \psi_{b[t-1]-1})}{c(e'')}$, then start a new batch, i.e., $b[t] = b[t-1]+1$,  observe the partial realization of all items selected so far, i.e., $\psi_{b[t]-1}=\psi_{b[t-1]-1}\cup\{(e, \Phi(e)) \mid e\in S_{[b[t-1]]}\}$. If $\max_{e\in  R\setminus \mathrm{dom}(\psi_{b[t]-1}) }\frac{\Delta(e\mid \mathrm{dom}(\psi_{b[t]-1}), \psi_{b[t]-1})}{c(e)} >0$ and adding $\arg\max_{e\in  R\setminus \mathrm{dom}(\psi_{b[t]-1}) }\frac{\Delta(e\mid \mathrm{dom}(\psi_{b[t]-1}), \psi_{b[t]-1})}{c(e)}$ to the solution does not violate the budget constraint, then add $\arg\max_{e\in  R\setminus \mathrm{dom}(\psi_{b[t]-1}) }\frac{\Delta(e\mid \mathrm{dom}(\psi_{b[t]-1}), \psi_{b[t]-1})}{c(e)}$ to the solution, and move to the next iteration, i.e., $t=t+1$; otherwise, terminate.
\end{itemize}
\end{itemize}

A detailed description of $\pi^k$ is presented in Algorithm \ref{alg:LPP2}.

\begin{algorithm}[hptb]
\caption{Equivalent $\pi^{k}$}
\label{alg:LPP3}
\begin{algorithmic}[1]
\STATE $t=1; b[0]=1; \psi_0=\emptyset; F=E; S_0=\emptyset; \forall i\in[n], S_{[i]}=\emptyset$.
\WHILE {$F\neq \emptyset$}
\STATE let $e' \leftarrow \arg\max_{e \in F} \frac{\Delta(e\mid  S_{t-1}, \psi_{b[t-1]-1})}{c(e)}$
\IF {$t=1$}
\STATE $e''\leftarrow e'$
\ENDIF
\IF {$\frac{\Delta(e'\mid  S_{t-1}, \psi_{b[t-1]-1})}{c(e')}\geq \alpha\cdot \frac{\Delta(e''\mid  \mathrm{dom}(\psi_{b[t-1]-1}), \psi_{b[t-1]-1})}{c(e'')}$}
\STATE \COMMENT{stay in the current batch}
\IF {$c(e')+c(S_{t-1})\leq B$}
\STATE consider $e'$
\STATE let $r \sim \mathrm{Bernoulli}(\frac{1}{2})$
\IF {$r=1$}
\STATE $b[t]=b[t-1]; e_t\leftarrow e'$; $S_t=S_{t-1}\cup\{e_t\}$; $S_{[b[t]]}=S_{[b[t]]}\cup\{e_t\}$; $F=F\setminus\{e_t\}; t=t+1$
\ELSE
\STATE $F=F\setminus\{e'\}$
\ENDIF
\ELSE
\STATE  break  \label{line:1}
\ENDIF
\ELSE
\STATE \COMMENT{start a new batch}
\STATE $b[t]=b[t-1]+1$
\STATE observe $\{(e, \Phi(e)) \mid e\in S_{[b[t-1]]}\}$; update $\psi_{b[t]-1}=\psi_{b[t-1]-1}\cup\{(e, \Phi(e)) \mid e\in S_{[b[t-1]]}\}$;
\STATE let $z \leftarrow \arg\max_{e\in F}\frac{\Delta(e\mid \mathrm{dom}(\psi_{b[t]-1}), \psi_{b[t]-1})}{c(e)}$
\IF {$\frac{\Delta(z \mid \mathrm{dom}(\psi_{b[t]-1}), \psi_{b[t]-1})}{c(e'')}>0$}
\IF {$c(z)+c(S_{t-1})\leq B$}
\STATE consider $z$
\STATE let $r \sim \mathrm{Bernoulli}(\frac{1}{2})$
\IF {$r=1$}
\STATE $e_t\leftarrow z$; $S_t=S_{t-1}\cup\{e_t\}$; $S_{[b[t]]}=S_{[b[t]]}\cup\{e_t\}$; $F=F\setminus\{e_t\}$; $e''\leftarrow z$; \COMMENT{$z$ is selected as the first item in batch $b[t]$}
\STATE $t=t+1$
\ELSE
\STATE $F=F\setminus\{z\}$
\ENDIF
\ELSE \STATE break \label{line:2} 
\ENDIF
\ELSE
\STATE break \label{line:3}
\ENDIF
\ENDIF
\ENDWHILE
\end{algorithmic}
\end{algorithm}

\subsection{Performance Analysis}
 For ease of analysis, we present an auxiliary policy in  Algorithm \ref{alg:LPP3}. Unlike Algorithm \ref{alg:LPP2} where $R$ is sampled at the beginning of the algorithm, we defer this decision in  Algorithm \ref{alg:LPP3}, that is, we toss a coin of success $1/2$ to decide whether or not to add an item to the solution  each time after an item is being considered. It is easy to verify that both versions of the algorithm have identical output distributions.

 We first provide some useful observations that will be used in the proof of the main theorem. Consider an arbitrary partial realization $\psi$, let $W(\psi)=\{e\in E\mid \Delta(e\mid \mathrm{dom}(\psi), \psi)>0\}$ denote the set of all items whose marginal utility with respect to $\mathrm{dom}(\psi)$ conditional on $\psi$ is positive. We number all items $e\in W(\psi)$  by decreasing ratio $\frac{\Delta(e\mid \mathrm{dom}(\psi), \psi)}{c(e)}$, i.e., $e(1)\in \arg\max_{e\in W(\psi)} \frac{\Delta(e\mid \mathrm{dom}(\psi), \psi)}{c(e)}$. If $\sum_{e\in W(\psi)} c(e)\geq B$, let $l=\min\{i\in \mathbb{N}\mid \sum_{j=1}^i c(e(i))\geq B\}$; otherwise, if $\sum_{e\in W(\psi)} c(e)< B$, let $l=|W(\psi)|$. Define $D(\psi)=\{e(i)\in W(\psi)\mid i\in[l]\}$ as the set containing the first $l$ items from $W(\psi)$. Intuitively, $D(\psi)$ represents a set of \emph{best-looking} items conditional on $\psi$.

Consider any $e\in D(\psi)$, assuming $e$ is the $i$-th item in $D(\psi)$, let
\begin{equation}
x(e, \psi)= \min\{1, \frac{B-\sum_{s \in \cup_{j\in[i-1]}\{e(j)\}} c(s)}{c(e)}\}~\nonumber
\end{equation}
where $ \cup_{j\in[i-1]}\{e(j)\}$ represents the first $i-1$ items in $D(\psi)$.

Define
\begin{eqnarray}
d(\psi) = \sum_{e\in D(\psi)} x(e, \psi)\Delta(e \mid \mathrm{dom}(\psi), \psi)~\nonumber
\end{eqnarray}

In analogy to Lemma 1 of \citep{gotovos2015non},
\begin{eqnarray}
\label{eq:ody}
d(\psi) \geq  \Delta(\pi^*\mid \mathrm{dom}(\psi), \psi)
\end{eqnarray}

%

We use $\lambda = (\{S^\lambda_t, \psi^\lambda_{b[t]-1}\mid t\in[z^\lambda]\}, \psi^\lambda_{b[z^\lambda]})$ to represent a fixed run of $\pi^{k}$ where $S^\lambda_t$ denotes the first $t$ items selected by $\pi^{k}$ under $\lambda$, $\psi^\lambda_{b[t]-1}$ denotes the partial realization of first $b[t]-1$ batches of selected items under $\lambda$, and $\psi^\lambda_{b[z^\lambda]}$ denotes the partial realization of all selected items under $\lambda$, i.e., $\pi^{k}$ selects $z^\lambda$ items under $\lambda$. Hence, $\mathrm{dom}(\psi^\lambda_{b[z^\lambda]}) = S^\lambda_{z^\lambda}$. Define $C(\lambda)$ as those items in $D(\psi^\lambda_{b[z^\lambda]})$ that have been considered by $\pi^{k}$ but not added to the solution because of the coin flips. Let $U(\lambda)$ denote those items in $D(\psi^\lambda_{b[z^\lambda]})$ that have not been considered by $\pi^{k}$. (\ref{eq:ody}) implies that
\begin{eqnarray}
d(\psi^\lambda_{b[z^\lambda]}) &=& \sum_{e\in D(\psi^\lambda_{b[z^\lambda]})  } x(e, \psi^\lambda_{b[z^\lambda]})\Delta(e \mid S^\lambda_{z^\lambda}, \psi^\lambda_{b[z^\lambda]}) \\
&=& \sum_{e\in U(\lambda) \cup C(\lambda)} x(e, \psi^\lambda_{b[z^\lambda]})\Delta(e \mid S^\lambda_{z^\lambda}, \psi^\lambda_{b[z^\lambda]})~\nonumber \\
&\geq & \Delta(\pi^*\mid S^\lambda_{z^\lambda}, \psi^\lambda_{b[z^\lambda]}) \label{eq:rainyagain}
\end{eqnarray}

Before presenting the main theorem, we provide two technical lemmas.
\begin{lemma}
\label{lem:bash}Let the random variable $\Lambda$ denote a random run of $\pi^{k}$,
 \begin{eqnarray}
 f_{avg}(\pi^{k}) \geq \frac{1}{2} \mathbb{E}[\sum_{e\in C(\Lambda)}\Delta(e\mid S^\Lambda_{z^\Lambda}, \psi^\Lambda_{b[z^\Lambda]})]
\end{eqnarray}
\end{lemma}
\emph{Proof:} For each $e\in E$, let $H(e) = \{\lambda \mid e\in C(\lambda)\}$ denote the set of all possible runs of $\pi^k$ under which $e$ is being considered and it is among \emph{best-looking} items. Let $\mathcal{D}(H(e))$ denote the  prior probability distribution over $H(e)$. In addition, let $H^+(e)$ denote the set of all possible runs of $\pi^k$ under which $e$ is being considered and let $\mathcal{D}(H^+(e))$ represent the  prior probability distribution over $H^+(e)$. Thus, $H(e) \subseteq H^+(e)$.  Consider any fixed run $\lambda \in H^+(e)$, assume $S^\lambda_e$ contains all items that are selected before $e$ is being considered and let $\psi^\lambda_e$ denote the partial realization of $S^\lambda_e$, i.e., $\mathrm{dom}(\psi^\lambda_e)=S^\lambda_e$. Note that $\psi^\lambda_e$  is a subrealization of $\psi^\lambda_{b[z^\lambda]}$, i.e.,  $\psi^\lambda_e \subseteq \psi^\lambda_{b[z^\lambda]}$. Then we have
\begin{eqnarray}
\label{eq:1112}
 f_{avg}(\pi^k)
  &=& \sum_{e\in E}\mathbb{E}_{\Lambda \sim \mathcal{D}(H^+(e))}[\Pr[e \mbox{ is selected given it was considered}]\Delta(e\mid S^\Lambda_e, \psi^\Lambda_e)]\\
  &\geq& \sum_{e\in E}\mathbb{E}_{\Lambda \sim \mathcal{D}(H(e))}[\Pr[e \mbox{ is selected given it was considered}]\Delta(e\mid S^\Lambda_e, \psi^\Lambda_e)]\\
 &=&  \sum_{e\in E}\mathbb{E}_{\Lambda \sim \mathcal{D}(H(e))}[\frac{1}{2}\times \Delta(e\mid S^\Lambda_e,  \psi^\Lambda_e)]\\
  &=& \frac{1}{2} \sum_{e\in E}\mathbb{E}_{\Lambda \sim \mathcal{D}(H(e))}[\Delta(e\mid  S^\Lambda_e, \psi^\Lambda_e)]\\
  &\geq& \frac{1}{2} \sum_{e\in E}\mathbb{E}_{\Lambda \sim \mathcal{D}(H(e))}[\Delta(e\mid S^\Lambda_{z^\Lambda}, \psi^\Lambda_{b[z^\Lambda]})]\\
  &=& \frac{1}{2} \mathbb{E}[\sum_{e\in C(\Lambda)}\Delta(e\mid S^\Lambda_{z^\Lambda}, \psi^\Lambda_{b[z^\Lambda]})]
\end{eqnarray}
The first inequality is due to  $H(e) \subseteq H^+(e)$. The second equality is due to the probability of $e$ being selected conditional on $e$ has been considered is $1/2$. The second inequality is due to $\psi^\lambda_e \subseteq \psi^\lambda_{b[z^\lambda]}$ for all $e\in E$ and $\lambda\in H(e)$ and $f : 2^{E\times O} \rightarrow \mathbb{R}_{\geq0}$ is adaptive submodular. $\Box$

\begin{lemma}
\label{lem:summer}Let the random variable $\Lambda$ denote a random run of $\pi^k$,
 \begin{eqnarray}
 f_{avg}(\pi^{k})+f(o)\geq\alpha\cdot \mathbb{E}_{\Lambda}[ \sum_{e\in U(\Lambda)}x(e, \psi^\Lambda_{b[z^\Lambda]})\Delta(e \mid S^\Lambda_{z^\Lambda}, \psi^\Lambda_{b[z^\Lambda]})]
\end{eqnarray}
\end{lemma}
\emph{Proof:} We first show that for any fixed run $\lambda$ of $\pi^k$, the following inequality holds:
\begin{eqnarray}
\label{eq:q}
\sum_{t\in[z^\lambda]} \Delta(e\mid S^\lambda_{t-1}, \psi^\lambda_{b[t]-1}) + f(o)\geq  \alpha\cdot \sum_{e\in U(\lambda)}x(e, \psi^\lambda_{b[z^\lambda]})\Delta(e \mid S^\lambda_{z^\lambda}, \psi^\lambda_{b[z^\lambda]})
\end{eqnarray}
We prove the above inequality in two cases, depending on whether Line \ref{line:1} or Line \ref{line:2} has been triggered or not under $\lambda$.
For the case when neither Line \ref{line:1} nor Line \ref{line:2} has been triggered, there are two possible reasons: either all items from $E$ have been considered or the \emph{best-looking} item with respect to $S^\lambda_{z^\lambda}$ conditional on $\psi^\lambda_{b[z^\lambda]}$ has negative marginal utility, i.e., Line \ref{line:3} is triggered. In the former case, (\ref{eq:q}) follows from the fact that $U(\lambda)=\emptyset$. In the latter case, due to the definition of $U(\lambda)$, it must hold that  for all $e\in U(\lambda)$, $\Delta(e \mid S^\lambda_{z^\lambda}, \psi^\lambda_{b[z^\lambda]})<0$, i.e., the marginal utility of every item in $U(\lambda)$ is negative. This implies that the RHS of (\ref{eq:q}) is non-positive, moreover, because $\sum_{t\in[z^\lambda]} \Delta(e\mid S^\lambda_{t-1}, \psi^\lambda_{b[t]-1})>0$ which implies that the LHS of (\ref{eq:q}) is non-negative, we have (\ref{eq:q}).

We next focus on proving (\ref{eq:q}) for the case when either Line \ref{line:1} or Line \ref{line:2} has been triggered. Let $e^\lambda_{z^\lambda+1}$ denote the item who triggers   Line \ref{line:1} or Line \ref{line:2}, i.e., adding  $e^\lambda_{z^\lambda+1}$ to $S^\lambda_{z^\lambda}$ violates the budget constraint. Let $b[\lambda_{z^\lambda+1}]=b[z^\lambda]$ if  Line \ref{line:1} has been triggered and let $b[\lambda_{z^\lambda+1}]=b[z^\lambda]+1$ if  Line \ref{line:2} has been triggered. 
 We first show that for any $t\in[z^\lambda+1]$,

 \begin{eqnarray}
\frac{\Delta(e^\lambda_t\mid S^\lambda_{t-1}, \psi^\lambda_{b[t]-1})}{c(e^\lambda_t)}
  &\geq& \alpha\cdot \max_{e\in U(\lambda)}\frac{\Delta(e \mid \mathrm{dom}(\psi^\lambda_{b[t]-1}), \psi^\lambda_{b[t]-1})}{c(e)}\\
   &\geq& \alpha\cdot \max_{e\in U(\lambda)}\frac{\Delta(e \mid S^\lambda_{z^\lambda}, \psi^\lambda_{b[z^\lambda]})}{c(e)}\label{eq:qq}
\end{eqnarray}
The first inequality is due to the selection rule of $\pi^{k}$ and the definition of $U(\lambda)$. The second inequality is due to $f: 2^{E \times O}\rightarrow \mathbb{R}_{\geq 0}$  is adaptive submodular and $ \psi^\lambda_{b[t]-1} \subseteq \psi^\lambda_{b[z^\lambda]}$ for any $t\in[z^\lambda+1]$.

Now we are ready to prove (\ref{eq:q}). Let $y\in \arg\max_{e\in U(\lambda)}\frac{\Delta(e \mid S^\lambda_{z^\lambda}, \psi^\lambda_{b[z^\lambda]})}{c(e)}$, (\ref{eq:qq}) implies that
\[\sum_{t\in[z^\lambda+1]}\Delta(e^\lambda_t\mid S^\lambda_{t-1}, \psi^\lambda_{b[t]-1}) \geq \alpha\cdot \sum_{t\in[z^\lambda+1]} \frac{c(e^\lambda_t) \Delta(y \mid S^\lambda_{z^\lambda}, \psi^\lambda_{b[z^\lambda]})}{c(y)}\] It follows that
 \begin{eqnarray}
\sum_{t\in[z^\lambda+1]}\Delta(e^\lambda_t\mid S^\lambda_{t-1}, \psi^\lambda_{b[t]-1}) &\geq& \alpha\cdot \sum_{t\in[z^\lambda+1]} \frac{c(e^\lambda_t) \Delta(y \mid S^\lambda_{z^\lambda}, \psi^\lambda_{b[z^\lambda]})}{c(y)}\\
&=&  \alpha\cdot (\sum_{t\in[z^\lambda+1]} c(e^\lambda_t))\frac{ \Delta(y \mid S^\lambda_{z^\lambda}, \psi^\lambda_{b[z^\lambda]})}{c(y)}\\
   &\geq&\alpha\cdot (\sum_{t\in[z^\lambda+1]} c(e^\lambda_t))\frac{ \sum_{e\in  U(\lambda)}x(e, \psi^\lambda_{b[z^\lambda]})\Delta(e \mid S^\lambda_{z^\lambda}, \psi^\lambda_{b[z^\lambda]})}{\sum_{e\in  U(\lambda)}x(e, \psi^\lambda_{b[z^\lambda]}) c(e)}\\
    &\geq&  \alpha\cdot \sum_{e\in U(\lambda)}x(e, \psi^\lambda_{b[z^\lambda]})\Delta(e \mid S^\lambda_{z^\lambda}, \psi^\lambda_{b[z^\lambda]})\label{eq:wang}
\end{eqnarray}

The last inequality is due to $\sum_{t\in[z^\lambda+1]} c(e^\lambda_t)=\sum_{e\in S^\lambda_{z^\lambda}}c(e)\geq B$ and $\sum_{e\in U(\lambda)}x(e, \psi^\lambda_{b[z^\lambda]})c(e)\leq B$. Recall that $o$ is the best singleton, we have $\sum_{t\in[z^\lambda]} \Delta(e\mid S^\lambda_{t-1}, \psi^\lambda_{b[t]-1}) + f(o) \geq \sum_{t\in[z^\lambda+1]}\Delta(e^\lambda_t\mid S^\lambda_{t-1}, \psi^\lambda_{b[t]-1})$ due to $f : 2^{E\times O} \rightarrow \mathbb{R}_{\geq0}$ is adaptive submodular. This together with (\ref{eq:wang}) implies (\ref{eq:q}).

Given (\ref{eq:q}) in hand, now we are ready to prove the lemma.
 \begin{eqnarray}
 f_{avg}(\pi^{k})+f(o)&=& \mathbb{E}_{\Lambda}[\sum_{t\in[z^\Lambda]} \Delta(e\mid S^\Lambda_{t-1}, \psi^\Lambda_{b[t]-1})+f(o)]\\
      &\geq&\mathbb{E}_{\Lambda}[ \alpha\cdot \sum_{e\in U(\Lambda)}x(e, \psi^\Lambda_{b[z^\Lambda]})\Delta(e \mid S^\Lambda_{z^\Lambda}, \psi^\Lambda_{b[z^\Lambda]})]\\
            &=&\alpha\cdot \mathbb{E}_{\Lambda}[  \sum_{e\in U(\Lambda)}x(e, \psi^\Lambda_{b[z^\Lambda]})\Delta(e \mid S^\Lambda_{z^\Lambda}, \psi^\Lambda_{b[z^\Lambda]})]
\end{eqnarray}
The inequality is due to (\ref{eq:q}). $\Box$

Now we are ready to present the main theorem.

\begin{theorem}
 If we randomly pick a policy from $\{o\}$ and $\pi^k$ with degree of adaptivity $\alpha$ to execute such that  $\{o\}$ is picked with probability $\frac{1/\alpha}{3+2/\alpha}$  and $\pi^k$ with degree of adaptivity $\alpha$ is picked with probability $\frac{3+ 1/\alpha}{3+2/\alpha}$, then we can achieve the expected utility of at least $\frac{1}{6+4/\alpha} f_{avg}(\pi^*)$.
\end{theorem}
\emph{Proof:} Lemma \ref{lem:bash} and Lemma \ref{lem:summer} imply that
 \begin{eqnarray}
 && 2 \times f_{avg}(\pi^{k}) + \frac{1}{\alpha}\times (f_{avg}(\pi^{k})+f(o)) \\
  &\geq&  \mathbb{E}_{\Lambda}[\sum_{e\in C(\Lambda)}\Delta(e\mid S^\Lambda_{z^\Lambda}, \psi^\Lambda_{b[z^\Lambda]})]+ \mathbb{E}_{\Lambda}[ \sum_{e\in U(\Lambda)}x(e, \psi^\Lambda_{b[z^\Lambda]})\Delta(e \mid S^\Lambda_{z^\Lambda}, \psi^\Lambda_{b[z^\Lambda]})]\\
  &=&  \mathbb{E}_{\Lambda}[\sum_{e\in C(\Lambda)}\Delta(e\mid S^\Lambda_{z^\Lambda}, \psi^\Lambda_{b[z^\Lambda]})+ \sum_{e\in U(\Lambda)}x(e, \psi^\Lambda_{b[z^\Lambda]})\Delta(e \mid S^\Lambda_{z^\Lambda}, \psi^\Lambda_{b[z^\Lambda]})]\\
  &\geq& \mathbb{E}_{\Lambda}[\sum_{e\in U(\Lambda) \cup C(\Lambda)} x(e, \psi^\Lambda_{b[z^\Lambda]})\Delta(e \mid S^\Lambda_{z^\Lambda}, \psi^\Lambda_{b[z^\Lambda]})]\\
  &\geq& \mathbb{E}_{\Lambda}[\Delta(\pi^*\mid S^\Lambda_{z^\Lambda}, \psi^\Lambda_{b[z^\Lambda]})]\\
  &=&f_{avg}(\pi^*@\pi^{k}) - f_{avg}(\pi^{k}) \label{eq:wolf}
\end{eqnarray}
The second inequality is due to $x(e, \psi^\lambda_{b[z^\lambda]})\leq 1$ for all $\lambda$. The third inequality is due to (\ref{eq:rainyagain}).

Recall that we toss a coin of success $1/2$ to decide whether or not to add an item to the
solution each time after an item is being considered, this design, together with Lemma 1 in  \citep{tang2021pointwise}, implies that  $f_{avg}(\pi^*@\pi^{k}) \geq f_{avg}(\pi^*)/2$. Combining this with  (\ref{eq:wolf}),
 \begin{eqnarray}
 2 \times f_{avg}(\pi^{k}) + \frac{1}{\alpha}\times (f_{avg}(\pi^{k})+f(o))  \geq  \frac{f_{avg}(\pi^*)}{2} - f_{avg}(\pi^{k})
\end{eqnarray}

It follows that
 \begin{eqnarray}
 \label{eq:cream}
(3+\frac{1}{\alpha})f_{avg}(\pi^{k})+\frac{1}{\alpha} f(o) \geq  \frac{f_{avg}(\pi^*)}{2}
\end{eqnarray}
Recall that our final policy randomly picks a policy from $\{o\}$ and $\pi^k$ to execute such that $\{o\}$ is picked with probability $\frac{1/\alpha}{3+2/\alpha}$  and $\pi^k$ is picked with probability $\frac{3+ 1/\alpha}{3+2/\alpha}$, then the expected utility of our final policy is
 \begin{eqnarray}
 \label{eq:pri}
f_{avg}(\pi^{k}) \times\frac{3+ 1/\alpha}{3+2/\alpha} +f(o)\times  \frac{1/\alpha}{3+2/\alpha}
\end{eqnarray}
(\ref{eq:cream}) and (\ref{eq:pri}) imply that the expected utility of our final policy is at least
 \begin{eqnarray}
 \label{eq:cream1}
\frac{(3+ 1/\alpha) f_{avg}(\pi^{k})+(1/\alpha) f(o)}{3+2/\alpha} \geq \frac{1}{6+4/\alpha} f_{avg}(\pi^*)
\end{eqnarray} $\Box$

\subsection{Bounding the Batch Query Complexity}
We next analyze the the batch query complexity, i.e., the number of batches a policy takes to complete the selection process,  of our partial-adaptive policy. We show that if we choose an appropriate $\alpha$, then it takes at most $\min\{O(n), O(\log n \log \frac{B}{c_{\min}})\}$, where $c_{\min}=\min_{e\in E} c(e)$ is the cost of the cheapest item, batches to achieve a constant approximation ratio. Notably, if we consider a cardinality constraint $k$, then the above complexity is upper bounded by $O(\log n \log k)$ which is polylogarithmic. This is in sharp contrast to the case of  fully adaptive policy whose batch query complexity is $O(n)$.  To establish this result, we first introduce two classes of stochastic functions.

\begin{definition}\citep{tang2021optimal}[Policywise Submodularity]
A function  $f: 2^{E\times O}\rightarrow \mathbb{R}_{\geq0}$ is policywise submodular with respect to a prior $p(\phi)$ and a knapsack constraint $(c, B)$ if for any two partial realizations $\psi$ and $\psi'$ such that $\psi'\subseteq \psi$ and $c(\mathrm{dom}(\psi)) \leq B$, and any $S\subseteq E$ such that $S\cap \mathrm{dom}(\psi) =\emptyset$, we have
\begin{eqnarray}
\max_{\pi\in \Omega}\Delta(\pi| \mathrm{dom}(\psi'), \psi') \geq \max_{\pi\in \Omega} \Delta(\pi| \mathrm{dom}(\psi), \psi)
\end{eqnarray}
where $\Omega = \{\pi\mid \forall \phi: c(E(\pi, \phi))\leq B-c(\mathrm{dom}(\psi)), E(\pi, \phi)\subseteq S \}$ denotes the set of feasible policies which are restricted to selecting items only from $S$.
\end{definition}

\begin{definition}[Weak Policywise Submodularity]
A function  $f: 2^{E\times O}\rightarrow \mathbb{R}_{\geq0}$ is weak policywise submodular with respect to a prior $p(\phi)$ and a knapsack constraint $(c, B)$ if for any partial realization $\psi$ such that $c(\mathrm{dom}(\psi)) \leq B$,
\begin{eqnarray}
f_{avg}(\pi^*) \geq \max_{\pi\in \Omega} \Delta(\pi| \mathrm{dom}(\psi), \psi)
\end{eqnarray}
where $\pi^*$ represents an optimal policy and $\Omega = \{\pi\mid \forall \phi: c(E(\pi, \phi))\leq B-c(\mathrm{dom}(\psi))\}$ denotes the set of feasible policies subject to a budget constraint $B-c(\mathrm{dom}(\psi))$.
\end{definition}

Observe that policywise submodularity implies weak policywise submodularity. In \cite{tang2021optimal}, it has been shown that policywise submodularity  (and hence weak policywise submodularity) can be found in a wide range of real-world applications, including generalized binary search \citep{golovin2011adaptive}, any applications where items are independent \citep{asadpour2016maximizing}, and adaptive viral marketing \citep{golovin2011adaptive}. We next show that if a utility function is both adaptive submodular and weak policywise submodular, then our partial-adaptive policy achieves a constant approximation ratio using only $O(\log n \log k)$ number of batches.

To analyze the batch query complexity of $\pi^k$, we focus on the original description of $\pi^k$ (Algorithm \ref{alg:LPP2}). That is, instead of tossing a coin each time after an item is being considered to be added to the solution, we first select a random subset $R$ which is obtained by including each item $e\in E$ independently with probability $1/2$. Then we run the ``density-greedy'' algorithm only on $R$. We first present a technical lemma that extends Lemma 7 in \citep{esfandiari2019adaptivity} to a general knapsack constraint.


\begin{lemma}
\label{lem:boston}
Consider an arbitrary batch index $l$ and a partial realization $\psi_l$ observed after the first $l$ batches  conditional on $R \subseteq E$ is being sampled at the beginning of $\pi^k$. Let $l^+ = l + \log_{\frac{1}{1-(1-\alpha)/2}}(\frac{n}{\delta})$. Assuming we run $\pi^k$ without budget constraint and $f: 2^{E \times O}\rightarrow \mathbb{R}_{\geq 0}$  is adaptive monotone and adaptive submodular, then
\begin{eqnarray}
\max_{e\in R\setminus \mathrm{dom}(\psi_{l^+})} \frac{\Delta(e\mid \mathrm{dom}(\psi_{l^+}), \psi_{l^+})}{c(e)} \leq (1-\frac{1-\alpha}{2}) \max_{e\in R\setminus \mathrm{dom}(\psi_l)} \frac{\Delta(e\mid \mathrm{dom}(\psi_l), \psi_l)}{c(e)}
\end{eqnarray} with probability at least $1-\delta$. Note that it could be the case that there may not be $l^+$ batches after selecting all items. If that is the case, just add some empty batches to make it $l^+$ batches.
\end{lemma}
\emph{Proof:} We closely follow the proof of Lemma 7 in \citep{esfandiari2019adaptivity}. For any $l' \geq l$, let the random variable $R_{l'}$ denote the set of items from $R\setminus \mathrm{dom}(\psi_{l'})$ such that $\frac{\Delta(e\mid \mathrm{dom}(\psi_{l'}), \psi_{l'})}{c(e)} \geq (1-\frac{1-\alpha}{2}) \max_{e\in R\setminus \mathrm{dom}(\psi_l)} \frac{\Delta(e\mid \mathrm{dom}(\psi_l), \psi_l)}{c(e)}$. To prove this lemma, it suffices to show that
\begin{eqnarray}
\label{eq:nips}
\mathbb{E}[|R_{l'+1}|] \leq (1-\frac{1-\alpha}{2})\mathbb{E}[|R_{l'}|]
\end{eqnarray}
 This is because (\ref{eq:nips}), together with the facts that $|R_l| \leq n$ and $l^+ = l + \log_{\frac{1}{1-(1-\alpha)/2}}(\frac{n}{\delta})$, implies that $\mathbb{E}[|R_{l^+}|] \leq \delta$. Note that $|R_{l^+}|$ is non-negative, thus we have $|R_{l^+}| = 0$ with probability at least $1-\delta$, as desired.

 We next focus on proving (\ref{eq:nips}). First, because of adaptive submodularity, we have $e\in R_{l'+1}$ implies that $e\in R_{l'}$. Thus, $R_{l'+1} \subseteq  R_{l'}$. We next show that for any item $e\in R_{l'}$, we have $e\notin R_{l'+1}$ with probability at least $\frac{1-\alpha}{2}$.

Because we run $\pi^k$ without budget constraint, if $e\in R_{l'}$ and $e$ is not selected within the $(l'+1)$-th batch of $\pi^k$, then
\begin{eqnarray}
\label{eq:nips?}
\mathbb{E}_{\Psi_{l'+1}}[\frac{\Delta(e\mid \mathrm{dom}(\Psi_{l'+1}), \psi_{l'})}{c(e)}] \leq  \alpha \max_{e\in R\setminus \mathrm{dom}(\psi_{l'})}\frac{\Delta(e\mid \mathrm{dom}(\psi_{l'}), \psi_{l'})}{c(e)}
 \end{eqnarray}
Because $\max_{e\in R\setminus \mathrm{dom}(\psi_{l'})}\frac{\Delta(e\mid \mathrm{dom}(\psi_{l'}), \psi_{l'})}{c(e)} \leq   \max_{e\in R\setminus \mathrm{dom}(\psi_l)} \frac{\Delta(e\mid \mathrm{dom}(\psi_l), \psi_l)}{c(e)}$ due to adaptive submodularity, we have $\mathbb{E}_{\Psi_{l'+1}}[\frac{\Delta(e\mid \mathrm{dom}(\Psi_{l'+1}), \psi_{l'})}{c(e)}] \leq \alpha  \max_{e\in R\setminus \mathrm{dom}(\psi_l)} \frac{\Delta(e\mid \mathrm{dom}(\psi_l), \psi_l)}{c(e)}$. This, together with the assumption that $f: 2^{E \times O}\rightarrow \mathbb{R}_{\geq 0}$  is adaptive monotone, implies that the probability that $\frac{\Delta(e\mid \mathrm{dom}(\Psi_{l'+1}), \psi_{l'})}{c(e)} < (1-\frac{1-\alpha}{2})  \max_{e\in R\setminus \mathrm{dom}(\psi_l)} \frac{\Delta(e\mid \mathrm{dom}(\psi_l), \psi_l)}{c(e)}$ is  at least $\frac{1-\alpha}{2}$. It follows that $e\notin R_{l'+1}$ with probability at least $\frac{1-\alpha}{2}$. This indicates that $\mathbb{E}[|R_{l'+1}|] \leq (1-\frac{1-\alpha}{2})\mathbb{E}[|R_{l'}|]$, as desired. $\Box$

 The following lemma shows that if $f: 2^{E \times O}\rightarrow \mathbb{R}_{\geq 0}$  is adaptive monotone and adaptive policywise submodular with respect to a prior $p(\phi)$ and any knapsack constraints, then it takes at most $O(\log n \log \frac{B}{c_{\min}})$ batches of $\pi^k$ to achieve a near optimal solution. Let  $\pi^k(T)$ denote a policy that runs $\pi^k$ and stops if it makes $T$ batch queries. In the rest of this section, let $T = \log_{\frac{1}{1-(1-\alpha)/2}}(\frac{n}{\delta})\times \log_{\frac{1}{1-(1-\alpha)/2}}(\frac{B}{c_{\min} (1-\alpha)})$  where $\delta=\frac{1-\alpha}{\log_{\frac{1}{1-(1-\alpha)/2}}(\frac{B}{c_{\min} (1-\alpha)})}$.
\begin{lemma}
\label{lem:main}
If  $f: 2^{E \times O}\rightarrow \mathbb{R}_{\geq 0}$  is adaptive monotone, adaptive submodular, and weak policywise submodular with respect to a prior $p(\phi)$ and any knapsack constraints, then $ f_{avg}(\pi^k(T)) \geq (1-\alpha + \alpha^2) f_{avg}(\pi^k) - (1-\alpha) f_{avg}(\pi^*)$.
\end{lemma}
\emph{Proof:} We first analyze the expected utility of $\pi^k(T)$ conditional on $R \subseteq E$ is sampled at the beginning of $\pi^k$. Let $a$ be the first
selected item, i.e., $a\in \arg\max_{e\in R} \frac{\Delta(e \mid \emptyset, \emptyset)}{c(e)}$. Assuming we run $\pi^k$ without budget constraint, by applying Lemma \ref{lem:boston} iteratively $\log_{\frac{1}{1-(1-\alpha)/2}}(\frac{B}{c_{\min} (1-\alpha)})$ times we have
\begin{eqnarray}
\label{eq:rainy}
\max_{e\notin \mathrm{dom}(\Psi_T)}\frac{\Delta(e \mid \mathrm{dom}(\Psi_T), \Psi_T)}{c(e)} &\leq& (1-\frac{1-\alpha}{2})^{\log_{\frac{1}{1-(1-\alpha)/2}}(\frac{B}{c_{\min} (1-\alpha)})}\frac{\Delta(a \mid \emptyset, \emptyset)}{c(a)} \\
&=&  \frac{c_{\min} (1-\alpha)}{B}\times \frac{\Delta(a \mid \emptyset, \emptyset)}{c(a)} \leq \frac{(1-\alpha)\Delta(a \mid \emptyset, \emptyset)}{B}
\end{eqnarray}
with probability $1-\delta \times \log_{\frac{1}{1-(1-\alpha)/2}}(\frac{B}{c_{\min} (1-\alpha)})=\alpha$. Recall that $T = \log_{\frac{1}{1-(1-\alpha)/2}}(\frac{n}{\delta})\times \log_{\frac{1}{1-(1-\alpha)/2}}(\frac{B}{c_{\min} (1-\alpha)})$, this indicates that with probability $\alpha$ the total expected marginal benefit of all item added after the $T$-th batch is at most
\begin{eqnarray}
\label{eq:rainy1}
\frac{(1-\alpha)\Delta(a \mid \emptyset, \emptyset)}{B} \times B = (1-\alpha)\Delta(a \mid \emptyset, \emptyset) \leq  (1-\alpha) \mathbb{E}[f_{avg}(\pi^k)\mid R]
\end{eqnarray}
where $\mathbb{E}[f_{avg}(\pi^k)\mid R]$ is the expected utility of $\pi^k$ conditional on $R$ is being sampled.

Moreover, because  $f: 2^{E\times O}\rightarrow \mathbb{R}$ is weak policywise submodular with respect to a prior $p(\phi)$ and any knapsack constraints, we have  \begin{eqnarray}
\label{eq:sunny}
\max_{\pi\in \Omega} \Delta(\pi | \mathrm{dom}(\psi_T), \psi_T) \leq f_{avg}(\pi^*)
 \end{eqnarray} where $\Omega = \{\pi\mid \forall \phi: c(E(\pi, \phi))\leq B-c(\mathrm{dom}(\psi_T))\}$ denotes the set of all feasible policies subject to a budget constraint $B-c(\mathrm{dom}(\psi_T))$. 
(\ref{eq:sunny}) indicates that for any $\psi_T$, the total expected marginal benefit of all item added after the $T$-th batch is at most $f_{avg}(\pi^*)$. This, together with (\ref{eq:rainy1}), implies that the total
expected marginal benefit of the items added after the $T$-th batch is at most
 \begin{eqnarray}\alpha (1-\alpha) \mathbb{E}[f_{avg}(\pi^k)\mid  R]+ (1-\alpha)  f_{avg}(\pi^*) 
 \end{eqnarray}





It follows that the expected utility of $\pi^k(T)$ conditional on $R$ is being sampled is at least
\begin{eqnarray}
\label{eq:parents}
&& \mathbb{E}[f_{avg}(\pi^k(T))\mid  R]  \geq f_{avg}(\pi^k) - \alpha(1-\alpha) \mathbb{E}[f_{avg}(\pi^k)\mid  R]- (1-\alpha)  f_{avg}(\pi^*) 
\end{eqnarray}

By taking the expectation of both sides of the above inequality over $R$, now we are ready to bound the expected utility of $\pi^k(T)$ as follows: 
 \begin{eqnarray}
 f_{avg}(\pi^k(T))&=& \sum_{R\subseteq E}\Pr[ R\mbox{ is sampled}]\mathbb{E}[f_{avg}(\pi^k(T))\mid  R] \\
      &\geq& \sum_{R\subseteq E}\Pr[ R\mbox{ is sampled}]( f_{avg}(\pi^k) - \alpha(1-\alpha) \mathbb{E}[f_{avg}(\pi^k)\mid  R]- (1-\alpha)  f_{avg}(\pi^*)) \\
       &=&  f_{avg}(\pi^k) -(1-\alpha) f_{avg}(\pi^*) -  \sum_{R\subseteq E}\Pr[ R\mbox{ is sampled}]\alpha(1-\alpha) \mathbb{E}[f_{avg}(\pi^k)\mid  R] \\
        &=&  f_{avg}(\pi^k) -(1-\alpha) f_{avg}(\pi^*) -  \alpha(1-\alpha) f_{avg}(\pi^k) \\
        &=& (1-\alpha + \alpha^2) f_{avg}(\pi^k) - (1-\alpha) f_{avg}(\pi^*)
\end{eqnarray}
The inequality is due to (\ref{eq:parents}). $\Box$

\emph{Remark:} Note that adaptive submodularity does not necessarily imply (\ref{eq:sunny}). As a result, one main result (Theorem 8) presented in \citep{esfandiari2019adaptivity} does not hold without resorting to weak policywise submodularity.

We next show that if we randomly pick a policy from $\pi^k(T)$ and $\{o\}$ to execute, then we can achieve a constant approximation ratio against the optimal fully adaptive policy.

\begin{lemma}
Assuming  $f: 2^{E \times O}\rightarrow \mathbb{R}_{\geq 0}$  is adaptive monotone, adaptive submodular, and adaptive policywise submodular with respect to a prior $p(\phi)$ and any knapsack constraints. If we randomly pick a policy from $\pi^k(T)$ and $\{o\}$ to execute such that $\pi^k(T)$ is picked with probability $\frac{3+ 1/\alpha}{3+2/\alpha}$ and $\{o\}$ is picked with probability  $\frac{1/\alpha}{3+2/\alpha}$, then we can achieve the expected utility of at least $\frac{(1-\alpha + \alpha^2)/2-(1-\alpha)(3+1/\alpha)}{3+2/\alpha}  f_{avg}(\pi^*)$.
\end{lemma}
\emph{Proof:}
Observe that if $\pi^k(T)$ is picked with probability $\frac{3+ 1/\alpha}{3+2/\alpha}$ and $\{o\}$ is picked with probability  $\frac{1/\alpha}{3+2/\alpha}$, then the expected utility can be represented as $f_{avg}(\pi^k(T)) \times \frac{3+ 1/\alpha}{3+2/\alpha} + f(o)\times  \frac{1/\alpha}{3+2/\alpha}$. Then we have
\begin{eqnarray}
 &&f_{avg}(\pi^k(T)) \times \frac{3+ 1/\alpha}{3+2/\alpha} + f(o)\times  \frac{1/\alpha}{3+2/\alpha} \\
 &\geq&  ((1-\alpha + \alpha^2) f_{avg}(\pi^k) - (1-\alpha) f_{avg}(\pi^*)) \times \frac{3+ 1/\alpha}{3+2/\alpha} + f(o)\times  \frac{1/\alpha}{3+2/\alpha}\\
 &\geq& \frac{(1-\alpha + \alpha^2)(3+ 1/\alpha)f_{avg}(\pi^k) +  (1/\alpha)f(o)}{3+2/\alpha} - (1-\alpha)\frac{3+ 1/\alpha}{3+2/\alpha} f_{avg}(\pi^*)\\
 &\geq& \frac{(1-\alpha + \alpha^2)((3+ 1/\alpha)f_{avg}(\pi^k) +  (1/\alpha)f(o))}{3+2/\alpha} - (1-\alpha)\frac{3+ 1/\alpha}{3+2/\alpha} f_{avg}(\pi^*)\\
  &\geq& \frac{(1-\alpha + \alpha^2)f_{avg}(\pi^*)/2}{3+2/\alpha}- (1-\alpha)\frac{3+ 1/\alpha}{3+2/\alpha} f_{avg}(\pi^*)\\
  &=&  \frac{(1-\alpha + \alpha^2)/2-(1-\alpha)(3+1/\alpha)}{3+2/\alpha}  f_{avg}(\pi^*)
\end{eqnarray}
The first inequality is due to Lemma \ref{lem:main}. The forth inequality is due to (\ref{eq:cream}). $\Box$
\section{Conclusion}
In this paper, we study the partial-adaptive submodular maximization problem. Our setting allows to select multiple items in a batch simultaneously and observe their realizations together. We develop effective solutions subject to both cardinality constraint and knapsack constraint. We analyze the batch query complexity of our policy under some additional assumptions about the utility function.
\bibliographystyle{ijocv081}
\bibliography{reference}

\begin{thebibliography}{14}
\expandafter\ifx\csname natexlab\endcsname\relax\def\natexlab#1{#1}\fi
\expandafter\ifx\csname url\endcsname\relax
  \def\url#1{{\tt #1}}\fi
\expandafter\ifx\csname urlprefix\endcsname\relax\def\urlprefix{URL }\fi
\expandafter\ifx\csname urlstyle\endcsname\relax
  \expandafter\ifx\csname doi\endcsname\relax
  \def\doi#1{doi:\discretionary{}{}{}#1}\fi \else
  \expandafter\ifx\csname doi\endcsname\relax
  \def\doi{doi:\discretionary{}{}{}\begingroup \urlstyle{rm}\Url}\fi \fi

\bibitem[{Amanatidis et~al.(2020)Amanatidis, Fusco, Lazos, Leonardi, and
  Reiffenh{\"a}user}]{amanatidis2020fast}
Amanatidis, Georgios, Federico Fusco, Philip Lazos, Stefano Leonardi, Rebecca
  Reiffenh{\"a}user. 2020.
\newblock Fast adaptive non-monotone submodular maximization subject to a
  knapsack constraint.
\newblock {\it Advances in neural information processing systems\/}.

\bibitem[{Asadpour and Nazerzadeh(2016)}]{asadpour2016maximizing}
Asadpour, Arash, Hamid Nazerzadeh. 2016.
\newblock Maximizing stochastic monotone submodular functions.
\newblock {\it Management Science\/} {\bf 62} 2374--2391.

\bibitem[{Badanidiyuru and Vondr{\'a}k(2014)}]{badanidiyuru2014fast}
Badanidiyuru, Ashwinkumar, Jan Vondr{\'a}k. 2014.
\newblock Fast algorithms for maximizing submodular functions.
\newblock {\it Proceedings of the twenty-fifth annual ACM-SIAM symposium on
  Discrete algorithms\/}. SIAM, 1497--1514.

\bibitem[{Buchbinder et~al.(2014)Buchbinder, Feldman, Naor, and
  Schwartz}]{buchbinder2014submodular}
Buchbinder, Niv, Moran Feldman, Joseph Naor, Roy Schwartz. 2014.
\newblock Submodular maximization with cardinality constraints.
\newblock {\it Proceedings of the twenty-fifth annual ACM-SIAM symposium on
  Discrete algorithms\/}. SIAM, 1433--1452.

\bibitem[{Chen and Krause(2013)}]{chen2013near}
Chen, Yuxin, Andreas Krause. 2013.
\newblock Near-optimal batch mode active learning and adaptive submodular
  optimization.
\newblock {\it ICML (1)\/} {\bf 28} 8--1.

\bibitem[{Esfandiari et~al.(2021)Esfandiari, Karbasi, and
  Mirrokni}]{esfandiari2019adaptivity}
Esfandiari, Hossein, Amin Karbasi, Vahab Mirrokni. 2021.
\newblock Adaptivity in adaptive submodularity.
\newblock {\it COLT\/} .

\bibitem[{Golovin and Krause(2011)}]{golovin2011adaptive}
Golovin, Daniel, Andreas Krause. 2011.
\newblock Adaptive submodularity: Theory and applications in active learning
  and stochastic optimization.
\newblock {\it Journal of Artificial Intelligence Research\/} {\bf 42}
  427--486.

\bibitem[{Gotovos et~al.(2015)Gotovos, Karbasi, and Krause}]{gotovos2015non}
Gotovos, Alkis, Amin Karbasi, Andreas Krause. 2015.
\newblock Non-monotone adaptive submodular maximization.
\newblock {\it Twenty-Fourth International Joint Conference on Artificial
  Intelligence\/}.

\bibitem[{Nemhauser et~al.(1978)Nemhauser, Wolsey, and
  Fisher}]{nemhauser1978analysis}
Nemhauser, George~L, Laurence~A Wolsey, Marshall~L Fisher. 1978.
\newblock An analysis of approximations for maximizing submodular set
  functions-i.
\newblock {\it Mathematical programming\/} {\bf 14} 265--294.

\bibitem[{Tang(2021{\natexlab{a}})}]{tang2021beyond}
Tang, Shaojie. 2021{\natexlab{a}}.
\newblock Beyond pointwise submodularity: Non-monotone adaptive submodular
  maximization in linear time.
\newblock {\it Theoretical Computer Science\/} {\bf 850} 249--261.

\bibitem[{Tang(2021{\natexlab{b}})}]{tang2021pointwise}
Tang, Shaojie. 2021{\natexlab{b}}.
\newblock Beyond pointwise submodularity: Non-monotone adaptive submodular
  maximization subject to knapsack and $k$-system constraints.
\newblock {\it Modelling, Computation and Optimization in Information Systems
  and Management Sciences\/}. Springer.

\bibitem[{Tang and Yuan(2020)}]{tang2020influence}
Tang, Shaojie, Jing Yuan. 2020.
\newblock Influence maximization with partial feedback.
\newblock {\it Operations Research Letters\/} {\bf 48} 24--28.

\bibitem[{Tang and Yuan(2021)}]{tang2021optimal}
Tang, Shaojie, Jing Yuan. 2021.
\newblock Optimal sampling gaps for adaptive submodular maximization.
\newblock {\it arXiv preprint arXiv:2104.01750\/} .

\bibitem[{Yuan and Tang(2017)}]{yuan2017no}
Yuan, Jing, Shaojie Tang. 2017.
\newblock No time to observe: adaptive influence maximization with partial
  feedback.
\newblock {\it Proceedings of the 26th International Joint Conference on
  Artificial Intelligence\/}. 3908--3914.

\end{thebibliography}




\end{document}